\documentclass[10pt,twocolumn,letterpaper]{article}

\usepackage{iccv}
\usepackage{times}
\usepackage{epsfig}
\usepackage{graphicx}
\usepackage{amsmath}
\usepackage{amssymb}

\usepackage{booktabs}
\usepackage{textcomp}
\usepackage{mathrsfs}
\usepackage[ruled,linesnumbered]{algorithm2e}
\usepackage[normalem]{ulem}
\usepackage{colortbl}
\usepackage{multirow}
\usepackage{booktabs}
\usepackage{graphicx}
\usepackage{hhline}

\usepackage{footmisc}
\usepackage{lipsum}

\usepackage[accsupp]{axessibility} 

\usepackage[pagebackref=true,breaklinks=true,letterpaper=true,colorlinks,bookmarks=false]{hyperref}

\iccvfinalcopy 
\def\iccvPaperID{11913} 

\ificcvfinal\fi

\begin{document}
\title{HopFIR: Hop-wise GraphFormer with Intragroup Joint Refinement for 3D Human Pose Estimation}
\author{
Kai Zhai\textsuperscript{1*}
\
\
\
Qiang Nie\textsuperscript{2*}
\
\
\
Bo Ouyang$^{1\dag}$
\
\
\
Xiang Li$^3$
\
\
\
Shanlin Yang$^1$
\\
{$^1$Hefei University of Technology}
\
\
\
{$^2$Youtu Lab, Tencent}
\
\
\
{$^3$Tsinghua University}
\\
}

\maketitle
\ificcvfinal\thispagestyle{empty}\fi

\newcommand\blfootnote[1]{%
\begingroup
\renewcommand\thefootnote{}\footnote{#1}%
\addtocounter{footnote}{-1}%
\endgroup
}

\blfootnote{*Equal Contribution}
\blfootnote{\dag Corresponding Author}

\begin{abstract}
    2D-to-3D human pose lifting is fundamental for 3D human pose estimation (HPE), for which graph convolutional networks (GCNs) have proven inherently suitable for modeling the human skeletal topology. However, the current GCN-based 3D HPE methods update the node features by aggregating their neighbors’ information without considering the interaction of joints in different joint synergies. Although some studies have proposed importing limb information to learn the movement patterns, the latent synergies among joints, such as maintaining balance are seldom investigated. We propose the Hop-wise GraphFormer with Intragroup Joint Refinement (HopFIR) architecture to tackle the 3D HPE problem. HopFIR mainly consists of a novel hop-wise GraphFormer (HGF) module and an intragroup joint refinement (IJR) module. The HGF module groups the joints by $k$-hop neighbors and applies a hop-wise transformer-like attention mechanism to these groups to discover latent joint synergies. The IJR module leverages the prior limb information for peripheral joint refinement. Extensive experimental results show that HopFIR outperforms the SOTA methods by a large margin, with a mean per-joint position error (MPJPE) on the Human3.6M dataset of 32.67 mm. We also demonstrate that the state-of-the-art GCN-based methods can benefit from the proposed hop-wise attention mechanism with a significant improvement in performance: SemGCN~\cite{zhao2019semantic} and MGCN~\cite{zou2021modulated} are improved by 8.9\% and 4.5\%, respectively.
\end{abstract}

\begin{figure}[t]
  \begin{center}
   \includegraphics[width=0.95\linewidth]{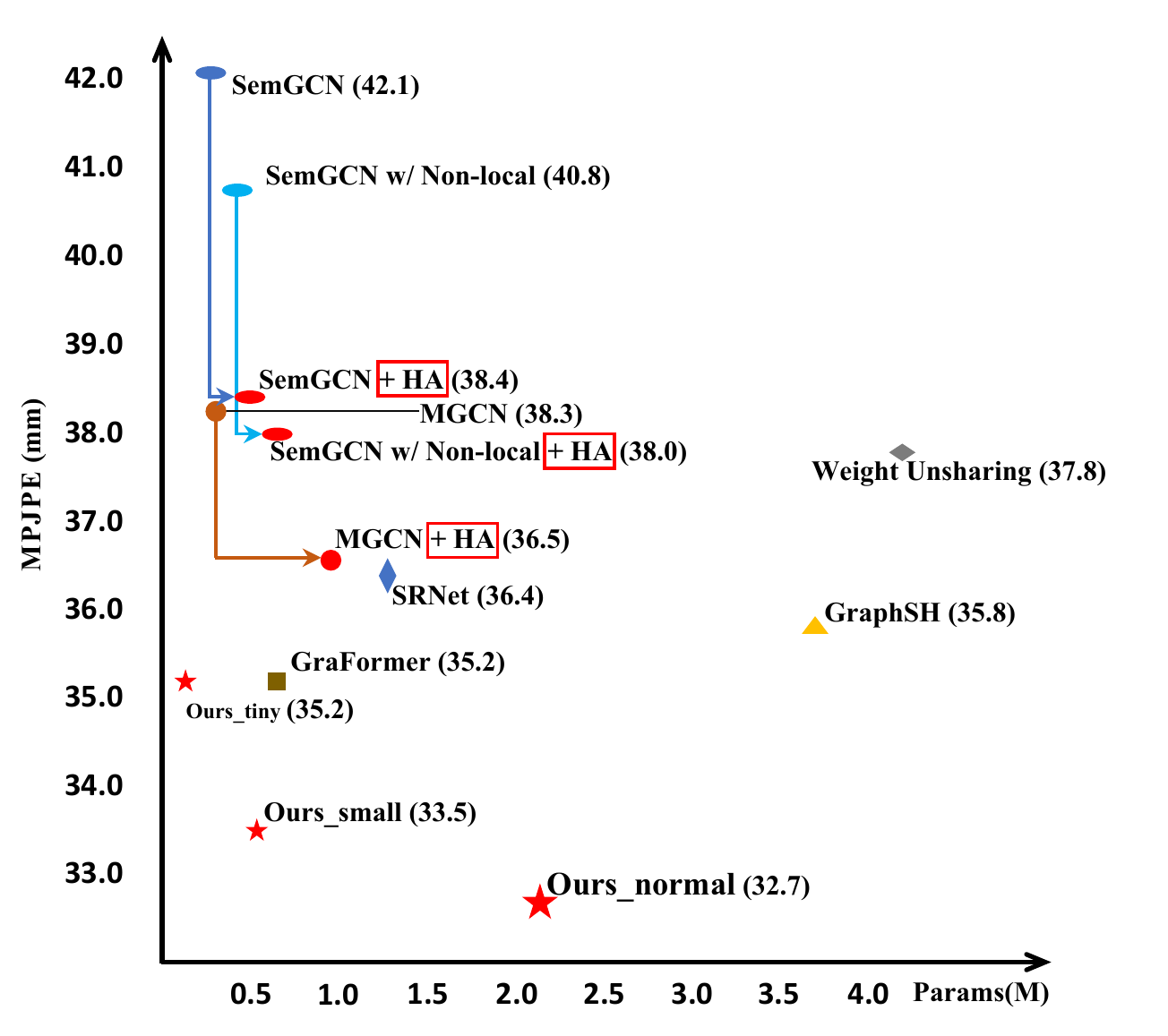}
   \end{center}
   \caption{  
   Comparison of performance and model size between the proposed HopFIR and SOTA methods, namely  Modulated GCN (MGCN)~\cite{zou2021modulated}, SemGCN~\cite{zhao2019semantic}, Weight Unsharing~\cite{liu2020comprehensive}, SRNet~\cite{zeng2020srnet}, GraphSH~\cite{xu2021graph},
   and GraFormer~\cite{zhao2022graformer}. The methods are evaluated on the Human3.6M dataset~\cite{ionescu2013h36m} with ground truth 2D joints as input.
   The arrow shows the performance improvement obtained by inserting the HA layer into the other networks. 
   Tiny, small, and normal denote the feature dimension of the HopFIR are 32, 64, and 128, respectively.
   }
   \label{fig:overview}
\end{figure}

\section{Introduction}

Monocular 3D human pose estimation aims to accurately regress the 3D locations of human joints in the camera coordinate system from a single image. It plays an important role in many applications, such as action recognition and human--computer interaction. Compared with the monocular systems, multi-view capture systems are expensive and inconvenient to set up and operate, which prevents them from being widely used in practice.
To tackle the monocular 3D HPE task, some approaches~\cite{cai2019exploiting, lee2018propagating, sun2017compositional, sun2018integral, yang20183d} estimates 3D joint coordinates or heat maps directly from an image via a convolutional neural network (CNN)~\cite{krizhevsky2017imagenet, lecun1998gradient}. However, direct regression from the image space suffers from the problem of a large parameter searching space, which always leads to a sub-optimal solution.
Recently, Martinez et al.~\cite{martinez2017simple} constructed a simple fully connected network using only 2D keypoints as input and achieved promising 3D HPE performance, showing that the 3D human pose can be efficiently and accurately estimated from 2D joint positions. 
Inspired by them and considering the thoroughly investigated 2D HPE, many works decompose the problem into two subtasks, i.e., the 2D HPE and 2D-to-3D pose lifting~\cite{liu2020comprehensive, zeng2021learning, zou2021modulated, xu2021graph}.
2D-to-3D pose lifting, therefore, has emerged as a fundamental task in this area that our work devotes to. 

A key consideration in 2D-to-3D pose lifting is that the human skeleton topology is inherently sparse and graph-structured. Fully connected neural networks are less effective in modeling graph-structured data due to their simple connections among all nodes and the probability of over-fitting. To leverage the information of the human skeletal topology, some works~\cite{xu2021graph, zeng2021learning, zou2021modulated} have proposed to model the human body with GCNs and have achieved SOTA results.
For example, Ci et al.~\cite{ci2019LCN} introduced a locally connected network to enhance the representation capability of the GCN, and Liu et al.~\cite{liu2020comprehensive} explored the weight sharing and feature transformation that occurs before or after feature aggregation in the GCN. 
One limitation of these GCN-based 3D HPE methods, however, is that they update the node features by aggregating their neighbors' information without considering the different contributions of these nodes to different joint synergies.

Instead of considering all the joints of a skeleton as a whole, Xue et al.~\cite{xue2022partaware} demonstrated that the human skeleton exhibits obvious part-wise inconsistency in its motion patterns, as also reported in SRNet~\cite{zeng2020srnet}. However, these works are limited to considering prior structural information of limb groupings and ignore investigating the latent groups underlying joint synergies. For example, the relative positions of the 1-hop neighbors of joint 0 are almost constant in ``Discussion" subject, which can be a latent group, as shown in Fig.~\ref{fig:group}.
Moreover, these works consider the joints in a limb as a whole to calculate the relationship with other limbs, which resulted in lower accuracy for peripheral joints, such as wrists and feet.

To address the abovementioned problems in monocular 3D HPE, we propose a novel architecture: the Hop-wise GraphFormer with Intragroup Joint Refinement (HopFIR).
The first key component of HopFIR is a novel hop-wise GraphFormer (HGF) module  that considers $k$-hop neighbors. In the HGF module, the information of every hop of every joint is aggregated into the hidden space, 
such that $N \times k$ groups of features are obtained for a skeleton model with $N$ joints.
Meanwhile, a hop-wise transformer-like attention mechanism is designed to extract the correlation among feature groups, which computes similarity by the dot product of the node feature and the group feature. The proposed HGF module enables the network to discover latent joint interactions considering human joint synergy. 
Because the HGF leverages little prior information about the human body and ignores the interaction among joints in a limb, especially the interactions of peripheral joints associated with a limb, we introduce an intragroup joint refinement (IJR) module to strengthen the intragroup correlation of joints grouped by limb prior information. 
Specifically, a residual block is built from two HGF modules followed by one IJR module. The proposed HopFIR architecture achieves optimal regression accuracy with a stack of three blocks.

To summarize, our work makes the following contributions:

$\bullet$ To the best of our knowledge, we design the first Hop-wise GraphFormer module to explore potential joint correlations underlying human joint synergy. We also prove that other GCN-based methods can benefit from the proposed HGF module efficiently, as shown in Fig.~\ref{fig:overview}.

$\bullet$ We design an Intragroup Joint Refinement module, which attends to intragroup joints to refine joint features through the associated limb, especially the wrists and feet. The IJR module enables HGF modules to discover the latent synergies among joints.

$\bullet$ We propose the novel Hop-wise GraphFormer with Intragroup Joint Refinement (HopFIR) architecture for 3D HPE, which is built entirely from HGF and IJR modules. Specifically, two HGF modules and one IJR module are coupled into a block.

$\bullet$ Extensive experiments demonstrate the effectiveness and generalizability of the proposed modules and HopFIR architecture by providing new state-of-the-art results on two challenging datasets, i.e., Human3.6M~\cite{ionescu2013h36m} and MPI-INF-3DHP~\cite{mehta20173dhp}.

\section{Related Work}
\label{sec:related work}

\textbf{3D Human Pose Estimation.} 
Early works~\cite{ramakrishna2012reconstructing, sminchisescu20083d} use handcrafted features, perspective relationships, and geometric constraints to estimate the 3D human pose. 
Recent pose estimation approaches can be generally divided into two categories. 
The first category of networks regresses 3D human joints directly from the image~\cite{pavlakos2017coarse, zhou2017towards}. Pavlakos et al.~\cite{pavlakos2017coarse} adopted a CNN to predict the voxel-wise likelihoods for each joint, and Zhou et al.~\cite{zhou2017towards} directly embedded a kinematic object model into the networks to learn the general multi-articulate object pose.
Approaches in the second category decouple the 3D HPE task into 2D pose estimation from an image and 3D pose estimation from the detected 2D joints (2D-to-3D). For example, Martinez et al.~\cite{martinez2017simple} proposed a simple yet effective baseline with fully-connect networks and proved that 3D human poses can be regressed simply and effectively from 2D keypoints. Our paper follows this pipeline and focuses on the 2D-to-3D pose lifting. 
For promoting 3D human pose regression accuracy, it is crucial to group joints with consideration of their interactions rather than treating all joints of a skeleton as a whole.
Xue et al.~\cite{xue2022partaware} divided the human skeleton graph into five groups according to the limbs to explore part-wise motion inconsistency, and Zeng et al.~\cite{zeng2020srnet} split the human joints into local regions and recombined the global information from the rest of the joints.
Our proposed HopFIR differs from these approaches by grouping joints by the $k$-hop neighbors of each joint and prior limb information, which enables the network to discover latent connections between groups in different human joint synergies.

\textbf{Graph Convolutional Networks.}
GCNs~\cite{defferrard2016convolutional, gori2005new, kipf2016semi, scarselli2008graph} generalize the capability of CNNs by performing convolution operations on graph-structured data. GCNs can be divided into two categories: the spectral-based approaches~\cite{defferrard2016convolutional} and the spatial-based approaches~\cite{kipf2016semi}.
Our approach falls into the second category, which applies message-passing operations on the graph nodes and their neighbors.

Due to the graph-structure topology of the human skeleton, many works~\cite{ci2019LCN, liu2020comprehensive} have introduced GCN to tackle the 3D HPE task. 
Zhao et al.~\cite{zhao2019semantic} proposed a SemGCN to learn the semantic relationships between human joints, and
Zou et al.~\cite{zou2021modulated} proposed a weight modulation and an affinity modulation based on the SemGCN.
These methods aggregate the first-order neighborhood messages to update the feature matrix by assigning different weights to different nodes. 
Some works~\cite{zhu2021posegtac, zou2020highorder} have extended the first-order neighbors to high-order neighbors in the spatial domain directly.
Zeng et al.~\cite{zeng2021learning} designed a hierarchical fusion block by dividing the fusion procedure into two stages, where all the high-order neighbors of a node are aggregated into a feature in the first stage and fuse it with the node feature and the first-order neighbor in the second stage.
Zhao et al.~\cite{zhao2022graformer} introduced Chebyshev graph convolution to fuse information among the $k$-hop neighbors of a joint directly.
These works updated features by aggregating each node's own $k$-hop neighborhood information in a GCN layer. However, HopFIR considers the $k$-hop groups of all nodes to reconstruct the $k$-hop feature of a node through the proposed attention mechanism, which can enhance the representation capability of GCNs. 

\textbf{Graph Attention.}
Graph attention networks~\cite{velivckovic2017gat} is a pioneer work that pay attention to the data in a graph structure by assigning an attention weight to each node. 
The introduction of the transformer~\cite{vaswani2017attention} for machine translation tasks has proven the capacity of attention for sequential input.
ViT~\cite{dosovitskiy2020vit} introduces the transformer into computer vision and achieves excellent performance. 
Inspired by them, some researchers have adopted the transformer for 3D HPE.
Zhao et al.~\cite{zhao2022graformer} applied self-attention to capture global information by calculating the similarity of all nodes. 
PoseFormer~\cite{zheng2021poseformer} directly applies Transformer Encoder (TE) by
viewing each joint and each frame as tokens in the spatial and temporal domains, respectively.
MixSTE~\cite{zhang2022mixste} is based on ~\cite{zheng2021poseformer}, where each joint feature is represented by the temporal TE to model the joint motion.
P-stmo~\cite{shan2022p} replaces the spatial TE with MLP and applies Stride Transformer Encoder to map $N$ frames to one frame. 
MHFormer~\cite{li2022mhformer} generates multi-hypothesis at different depths of stacked spatial TEs by viewing all frames of each joint feature as a token, then communicates these hypotheses with cross-attention and self-attention.
Different from the existing graph attention for 3D HPE, we propose the intergroup multi-head attention mechanism among the $k$-hop groups of all nodes, which assigns the attention weights by computing the similarity between the node feature and $k$-hop group feature. Moreover, we introduce the intragroup multi-head self-attention in limb groups to refine the joint features and promote the HGF module to discover the latent synergies among joints.

\section{The Proposed HopFIR}
\label{sec:formatting}

This paper proposes a novel architecture to regress the 3D human pose from $N$ given 2D keypoints $X\in\mathbb{R}^{N\times2}$. The proposed framework mainly consists of the HGF and IJR modules. In this section, we first review the vanilla Graph Convolutional Network and Transformer in Sec.~\ref{subsec:vanilla}. We then introduce the HGF and IJR modules in Sec.~\ref{subsec:HGF} and Sec.~\ref{subsec:IJR}, respectively. Finally, we present the network architecture in Sec.~\ref{subsec:architecture}.

\begin{figure}[t]
  \begin{center}
   \includegraphics[width=0.9\linewidth]{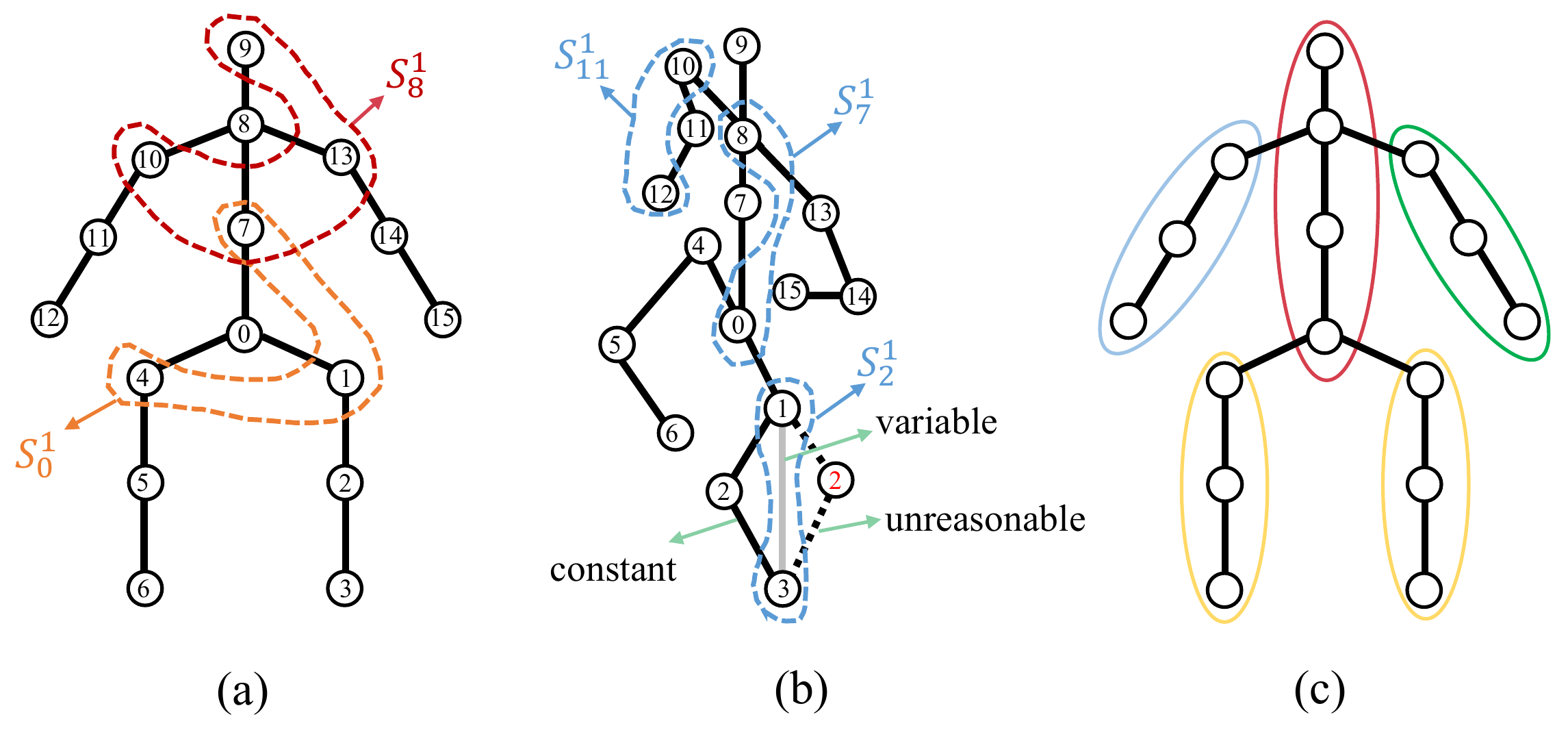}
   \end{center}
   \caption{An illustration of the human skeleton graph and the groups in HGF and IJR modules. The $k$-hop neighbors of a joint are set as a group in HGF modules where (a) indicates the 1-hop groups of different joints and (b) indicates several latent groups because of physical limitations. (c) indicates the joints grouped by prior limb information.
   }
   \label{fig:group}
\end{figure}

\begin{figure*}[t]
  \begin{center}
  \includegraphics[width=0.8\linewidth]{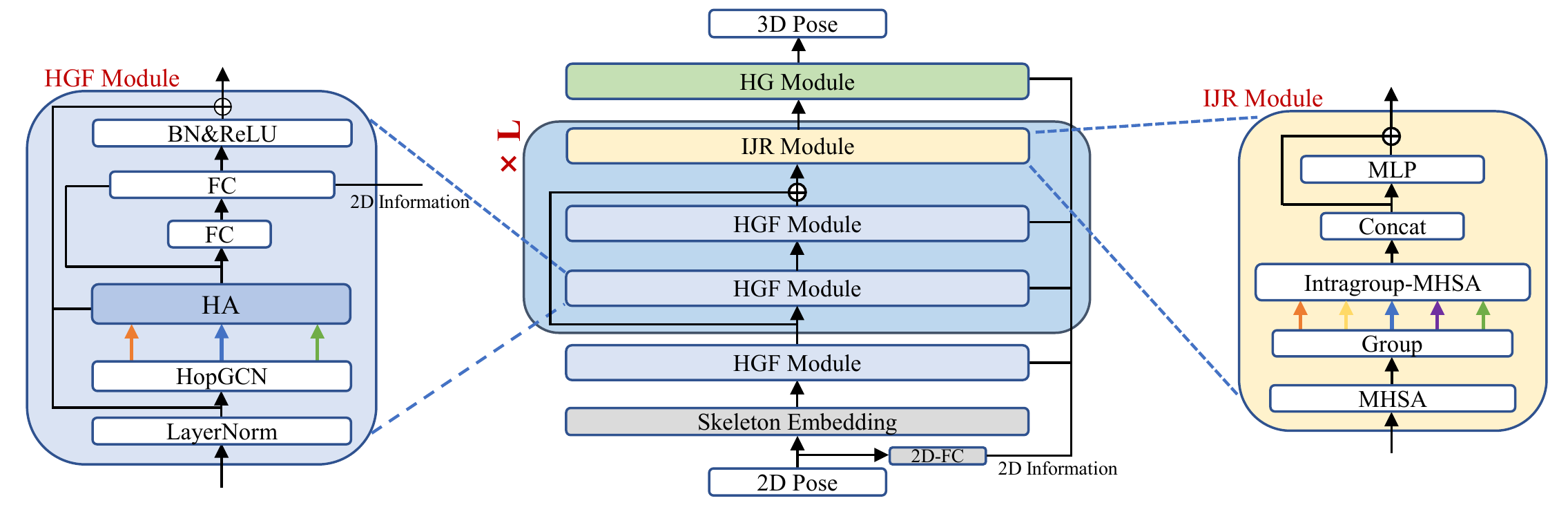}
  \end{center}
  \caption{The HopFIR architecture, with details of the HGF and IJR modules. 
  The designed block is a residual block~\cite{he2016resnet} built by two HGF modules followed by one IJR module. The proposed architecture with three blocks achieves optimal performance. Arrows of different colors represent different hops and groups in HGF and IJR, respectively.
  }
  \label{fig:framework}
\end{figure*}

\subsection{Vanilla GCN and Transformer}
\label{subsec:vanilla}

\textbf{GCN.} A graph is defined as $\mathcal{G} =(\mathcal{V}, \mathcal{E})$, where $\mathcal{V}$ is a set of $N$ nodes and $\mathcal{E}$ is the adjacency matrix representing the edges between the nodes. Given a collection of input features $H^l \in \mathbb{R}^{N \times D}$,
a generic GCN layer that aggregates neighborhood information can be formulated as follows:
\begin{equation}
\label{e1}
    H^{(l+1)}=\sigma(\tilde{A} H^l W)
\end{equation}
where $W\in\mathbb{R}^{D \times D'}$is the learnable weight matrix that transforms the feature dimension from $D$ to $D'$, 
and $\sigma(.)$ is the activation function, such as ReLU~\cite{nair2010relu}. $H^{(l+1)}$ is the updated feature matrix, 
$\tilde{A}\in \mathbb{R}^{N\times N}$ is the symmetrically normalized affinity matrix~\cite{kipf2016semi} with added self-connections, and
$A \in \{0,1\}^{N \times N}$ is the adjacency matrix. The $(i, j)th$ entry $a_{ij}=1$ representing node $j$ is the neighbor of node $i$. Otherwise, they are not connected and $a_{ij}=0$. Therefore, the none-neighbor nodes have a weak influence on each other in the vanilla GCN, which hinders the modeling of underlying joint synergy in 3D HPE. 

\textbf{Transformer.} Transformer architecture relies entirely on self-attention to compute representations of its input and output. The self-attention function maps the inputs to the queries $Q$, keys $K$, and values $V$ by weight matrices $W_Q$, $W_K$, and $W_V$, respectively, and the matrix of outputs is calculated as:
\begin{equation}
    Attention(Q, K, V)=Softmax(QK^T/\sqrt{d})V
\end{equation}
where $d$ is the feature dimension of $Q$, and $\frac{1}{\sqrt{d}}$ is a scaling factor to prevent extremely small gradients.
The multi-head self-attention (MHSA), which performs self-attention in parallel, projects the queries, keys, and values $P$ times with different linear projections to the respective subspaces, as follows:
\begin{equation}
    MultiHead(Q, K, V) = Concat(D_1,...,D_P)W_{o} 
\end{equation}
where $W_O$ is the projection matrix of outputs, $D_p = Attention(QW_Q^p, KW_K^p, VW_V^p)$, and $p\in [1,..,P].$ 

\subsection{Hop-wise GraphFormer}
\label{subsec:HGF}
Previous GCN studies for 3D HPE aggregate multi-hop neighborhood information~\cite{zeng2021learning, zou2020highorder} or assign an attention weight to each first-order neighbor~\cite{zhao2019semantic,zou2021modulated}. To effectively capture the node's neighborhood message and increase the representational capacity of GCNs, we introduce the hop-wise GraphFormer (HGF) module, which treats each hop as a group (Fig.\ref{fig:group} a,b) and computes the attention weights for each hop (more intuitive descriptions for $k$-hop can be found in the supplementary material). By considering the relationship within $k$ hops, we can obtain $N \times k$ groups for a skeleton graph with $N$ joints, which provides enough combinations of joints to discover the latent correlations among joints in different human joint synergy.

We first define the $k$-hop matrix $A^k$ as
\begin{equation}
    a^k_{ij} = \begin{cases}
    1, &d(v_i, v_j)=k\\
    0, &otherwise
    \end{cases}
\end{equation}
where $d(v_i, v_j)$ denotes the distance of the shortest path between $v_i$ and $v_j$ on the skeleton graph.
The $k$-hop neighborhood information is aggregated with a weighted sum of the target node's $k$-hop neighbors, named HopGCN:
\begin{equation}
\label{e5}
    s^k_i = \sum_{j}a^k_{ij}h_jW^k
\end{equation}
where $s^k_i \in \mathbb{R}^{D}$ is a hidden representation of the $k$-hop neighborhood information.
Eq.~\ref{e5} is similar to Eq.~\ref{e1} but with the extended definition of $A$ to $k$ hops. The weight matrix $W$ is assigned to the respective hops. 

Before aggregating the hidden representation $s^k_i$ to the target node, we propose a transformer-like attention mechanism computing the similarity between the node feature $h_i$ and the $k$-hop hidden representation, as shown in Eq.~\ref{e7}:
\begin{equation}
\label{e7}
    z^k_i=\sum_jsoftmax_j(\frac{h_{i}{s^k_j}^T}{\sqrt{d}})s^k_j.
\end{equation}
Due to the structure of the human skeleton and the joint synergy, the $k$-hop neighborhood of joint $i$ is related to that of other joints. In Fig.~\ref{fig:group}, for example, the synergy of hop $S_5^1$ is related to hop $S_6^1$ and $S_3^1$. By packing $z^k_i$, $h_i$, and $s^k_i$ into matrices $Z^k$, $H$, and $S^k$, we can calculate $Z^k$ in parallel:
\begin{equation}
    Z^k=softmax(\frac{H{S^k}^T}{\sqrt{d}})S^k
\end{equation}
$Z^k$ can also be reconstructed using the self-attention mechanism purely on hop features or swap the positions of $H$ and $S^k$. A performance comparison of these cases is discussed in the experimental results.
 
We further reduce the dimension of hidden representation by using a fully connected(FC) layer based on the order of the neighborhood, considering different amounts of information related to the target node, as formulated in Eq.~\ref{eq:fc}.
\begin{equation}\label{eq:fc}
    r^k_i=F^k(z^k_i)
\end{equation}
where $r^k_i$ is the refined representation of the $k$-hop neighborhood information with respect to node $i$, and $F^k$ is the mapping function. 
The refined representation is then concatenated to the updated node feature. 

All the refined representations of hop-wise neighborhoods and the target node feature $h_i$ are encoded to a $D$-dimensional vector:
\begin{equation}
\label{e10}
    h'_{i}=F(h_i, r_i^1, r_i^2, r_i^3, ..., r_i^k)  
\end{equation}
where $h'_i$ is the final updated feature of node $i$ in this layer, and $F$ is the aggregation function. 
Packing together the updated features of all nodes, Eq.~\ref{e10} can be rewritten as:
\begin{equation}
    H' = F(H, R^1, R^2, R^3, ..., R^k).
\end{equation}
We thereby obtain as the core layer of the HGF module the hop-wise attention (HA) layer, which extracts latent correlations between feature groups and aggregates $k$-hop neighborhood information.
With the increment of $k$, we get more $N$ groups on which to explore the underlying joint synergies.
The value of $k$ can be adjusted based on the tasks and pipelines. According to the experimental results, three hops achieve optimal performance in the HopFIR architecture.

To fuse the current global information and original 2D information, we concatenate all the joint features in a batch and feed them into a FC layer to extract the current global information. We then concatenate the extracted global information, 2D information, and the output of the HA layer on the feature dimension of the joint feature to fuse all the information in another FC layer.

\subsection{Intragroup Joint Refinement}
\label{subsec:IJR}
The proposed HGF module splits the skeleton graph into different groups based on the $k$-hop neighborhood of each joint, which attends to the key groups in joint synergies.
However, HGF leverages little prior information about the human body and ignores the interaction among joints in a limb, especially the interaction of the peripheral joints associated with limbs, such as the wrists and feet. 
We introduce an intragroup joint refinement (IJR) module to strengthen the intragroup correlation of joints grouped by limb prior information, as shown in Fig.~\ref{fig:group}(c). The HGF features of the joints in each limb group are refined in the IJR module by using multi-head self-attention~\cite{vaswani2017attention}, as in Eq.~\ref{eq:mhsa}.
\begin{equation}
\label{eq:mhsa}
    H_g = MHSA(MHSA(H)_g)
\end{equation}
where $H$ is the feature matrix from the HGF module, $MHSA(H)_g$ is the feature matrix of the group $g$ updated by global multi-head self-attention, and $H_g$ is the final feature matrix of the group $g$ updated by the IJR module. More details of the IJR module are provided in Fig.~\ref{fig:framework}.

\subsection{The HopFIR Architecture}
\label{subsec:architecture}
The HopFIR architecture consists of the proposed HGF and IJR modules, as illustrated in Fig.~\ref{fig:framework}. The residual block, which contains two HGF modules and one IJR module, is designed as the basic block in HopFIR.
Moreover, we define a linear embedding layer to map the input to the latent space and a HG module to transform the output into 3D space. 
The HG module is a variant of HGF designed for the output layer, more details of which are provided in the supplementary material.
HopFIR accepts 2D keypoints as input, which can be obtained via an off-the-shelf 2D detector.
The graph is obtained by adding a normalized globally learnable $k$-hop graph to the skeleton graph and then symmetrically normalizing it, as in~\cite{zou2021modulated}.
We use the L1-norm loss and L2-norm loss to compute the error between ground truth and prediction with the weighted sum as follows:
\begin{equation}
    L = \alpha \sum_{n=1}^N \Vert Y_n - \hat{Y_n} \Vert_2 + \beta \sum_{n=1}^N \Vert Y_n - \hat{Y_n} \Vert_1
\end{equation}
where $N$ is the joint number, $\hat{Y_n}$ is the predicted 3D position of joint $n$, $Y_n$ is the ground truth, $\alpha$ = 1, and $\beta$ = 0.1.

\begin{table*}[t]
    \begin{center}
    \setlength{\extrarowheight}{0pt}
    \setlength{\aboverulesep}{0pt}
    \setlength{\belowrulesep}{0pt}
    \resizebox{\linewidth}{!}{
    \begin{tabular}{l@{ }l| c c c c c c c c c c c c c c c c}
        \toprule
        \textbf{Protocol \#1}& & Dir. & Disc. & Eat & Greet & Phone & Photo & Pose & Pur. & Sit & SitD. & Smoke & Wait & WalkD. & Walk & WalkT. & Avg.\\
        \midrule\midrule
        Martinez et al.~\cite{martinez2017simple} &ICCV2017& 51.8&  56.2&  58.1&  59.0& 69.5 & 78.4 & 55.2 & 58.1 & 74.0 & 94.6 & 62.3 & 59.1 & 65.1 & 49.5 & 52.4 & 62.9\\ 
        Zhao et al.~\cite{zhao2019semantic}($\dagger$) &CVPR2019& 47.3 & 60.7 & 51.4 & 60.5 & 61.1 &\textbf{49.9} &\underline{47.3} & 68.1 & 86.2 &\textbf{55.0} & 67.8 & 61.0 &\textbf{42.1} & 60.6 & 45.3 & 57.6\\
        Ci et al.~\cite{ci2019LCN}($\dagger$) &ICCV2019& 46.8 & 52.3 &\textbf{44.7} & 50.4 & 52.9 & 68.9 & 49.6 & 46.4 & 60.2 & 78.9 & 51.2 & 50.0 & 54.8 & 40.4 & 43.3 & 52.7\\
        Pavllo et al.~\cite{pavllo20193d} &CVPR2019&47.1 &50.6 &49.0 &51.8 &53.6 &61.4 &49.4& 47.4 &59.3& 67.4 &52.4 &49.5& 55.3 &39.5& 42.7 &51.8\\
        Cai et al.~\cite{cai2019exploiting}($\dagger$) &ICCV2019&46.5 &48.8 &47.6 &50.9& 52.9& 61.3& 48.3& 45.8& 59.2& 64.4 &51.2 &48.4 &53.5 &39.2 &41.2 &50.6\\
        Liu et al.~\cite{liu2020comprehensive}($\dagger$) &ECCV2020& 46.3 & 52.2 & 47.3 & 50.7 & 55.5 & 67.1 & 49.2 & 46.0 & 60.4 & 71.1 & 51.5 & 50.1 & 54.5 & 40.3 & 43.7 & 52.4\\
        zeng et al.~\cite{zeng2020srnet} &ECCV2020&\underline{44.5} &\underline{48.2} & 47.1 & \textbf{47.8} & 51.2 &\underline{56.8} & 50.1 &\underline{45.6} & 59.9 & 66.4 & 52.1 &\textbf{45.3} & 54.2 & 39.1 &\underline{40.3} & 49.9\\ 
        Zou et al.~\cite{zou2021modulated}($\dagger$) &ICCV2021& 45.4 & 49.2 & 45.7 & 49.4 &\underline{50.4} & 58.2 & 47.9 & 46.0 &\underline{57.5} & 63.0 &\underline{49.7} & 46.6 & 52.2 &\underline{38.9} & 40.8 &\underline{49.4}\\ 
        Xu et al.~\cite{xu2021graph}($\dagger$) &CVPR2021& 45.2 & 49.9 & 47.5 & 50.9 & 54.9 & 66.1 & 48.5 & 46.3 & 59.7 & 71.5 & 51.4 & 48.6 & 53.9 & 39.9 & 44.1 & 51.9\\
        Zhao et al.~\cite{zhao2022graformer} ($\Delta$)($\dagger$) &CVPR2022& 45.2 & 50.8 & 48.0 & 50.0 & 54.9 & 65.0 & 48.2 & 47.1 & 60.2 & 70.0 & 51.6 & 48.7 & 54.1 & 39.7 & 43.1 & 51.8\\
        \midrule
        Ours($\Delta$)($\dagger$)& &\textbf{43.9} &\textbf{47.6} &\underline{45.5} &\underline{48.9} &\textbf{50.1} &58.0 &\textbf{46.2} &\textbf{44.5} &\textbf{55.7} &\underline{62.9} &\textbf{49.0} &\underline{45.8} &\underline{51.8} &\textbf{38.0} &\textbf{39.9} &\textbf{48.5}\\
        \bottomrule
    \end{tabular}
    }
    \end{center}
    \caption{
    Quantitative comparison on Human3.6M with detected 2D poses as input under Protocol \#1, in millimeters. The best results are highlighted in bold and the second-best results are underlined. ($\dagger$) indicates GCN-based methods and ($\Delta$) indicates Transformer-based methods.
    }
    \label{table:2D}
\end{table*}

\begin{table*}[ht]
    \begin{center}
    \setlength{\extrarowheight}{0pt}
    \setlength{\aboverulesep}{0pt}
    \setlength{\belowrulesep}{0pt}
    \resizebox{\linewidth}{!}{
    \begin{tabular}{l@{ }l|c c c c c c c c c c c c c c c c}
        \toprule
        \textbf{Protocol \#1} && Dir. & Disc. & Eat & Greet & Phone & Photo & Pose & Pur. & Sit & SitD. & Smoke & Wait & WalkD. & Walk & WalkT. & Avg.\\
        \midrule\midrule
        Zhou et al.~\cite{zhou2019hemlets}(+) &ICCV2019& 34.4 & 42.4 & 36.6 & 42.1 & 38.2 & \textbf{39.8} & 34.7 & 40.2 & 45.6 & 60.8 & 39.0 & 42.6 & 42.0 & 29.8 & 31.7 & 39.9\\
        Ci et al.~\cite{ci2019LCN}(+)($\ast$)($\dagger$) &ICCV2019& 36.3 & 38.8 & 29.7 & 37.8 & 34.6 & 42.5 & 39.8 & 32.5 & 36.2 & \textbf{39.5} & 34.4 & 38.4 & 38.2 & 31.3 & 34.2 & 36.3\\
        \midrule
        Martinez et al.~\cite{martinez2017simple} &ICCV2017& 37.7 & 44.4 & 40.3 & 42.1 & 48.2 & 54.9 & 44.4 & 42.1 & 54.6 & 58.0 & 45.1 & 46.4 & 47.6 & 36.4 & 40.4 & 45.5\\ 
        Zhao et al.~\cite{zhao2019semantic}($\dagger$)&CVPR2019& 37.8 & 49.4 & 37.6 & 40.9 & 45.1 & 41.4 & 40.1 & 48.3 & 50.1 & 42.2 & 53.5 & 44.3 & 40.5 & 47.3 & 39.0 & 43.8\\
        Cai et al.~\cite{cai2019exploiting}($\dagger$) &ICCV2019& 33.4 & 39.0 & 33.8 & 37.0 & 38.1 & 47.3 & 39.5 & 37.3 & 43.2 & 46.2 & 37.7 & 38.0 & 38.6 & 30.4 & 32.1 & 38.1\\
        Liu et al.~\cite{liu2020comprehensive}($\dagger$) &ECCV2020& 36.8 & 40.3 & 33.0 & 36.3 & 37.5 & 45.0 & 39.7 & 34.9 & 40.3 & 47.7 & 37.4 & 38.5 & 38.6 & 29.6 & 32.0 & 37.8\\
        zeng et al.~\cite{zeng2020srnet} &ECCV2020& 35.9 & 36.7 & 29.3 & 34.5 & 36.0 & 42.8 & 37.7 & 31.7 & 40.1 & 44.3 & 35.8 & 37.2 & 36.2 & 33.7 & 34.0 & 36.4\\
        Zou et al.~\cite{zou2021modulated}($\dagger$) &ICCV2021& - & - & - & - & - & - & - & - & - & - & - & - & - & - & - & 37.4\\ 
        Xu et al.~\cite{xu2021graph}($\dagger$) &CVPR2021& 35.8 & 38.1 & 31.0 & 35.3 & 35.8 & 43.2 & 37.3 & 31.7 & 38.4 & 45.5 & 35.4 & 36.7 & 36.8 & 27.9 & 30.7 & 35.8\\
        Zhao et al.~\cite{zhao2022graformer} ($\Delta$)($\dagger$)&CVPR2022& 32.0 & 38.0 & 30.4 & 34.4 & 34.7 & 43.3 & 35.2 & 31.4 & 38.0 & 46.2 & 34.2 & 35.7 & 36.1 & 27.4 & 30.6 & 35.2\\
        \midrule
        Ours($\Delta$)($\dagger$)& &\textbf{31.3} & \textbf{34.0} & \textbf{28.0} & \textbf{32.0} & \textbf{33.1} & 42.1 & \textbf{34.1} & \textbf{28.1} & \textbf{33.6} & 39.8 & \textbf{31.7} & \textbf{32.9} & \textbf{33.8} & \textbf{26.7} & \textbf{28.9} & \textbf{32.7}\\ 
        \bottomrule
    \end{tabular}
    }
    \end{center}
    \caption{
    Quantitative comparison on Human3.6M with ground truth 2D keypoints as input under Protocol \#1, in millimeters. (+) uses additional data from MPII~\cite{andriluka2014mpii}. ($\ast$) uses pose scales in both training and testing.
    The best results are highlighted in bold.}
    \label{table:GT1}
\end{table*}

\section{Experiments}
In this section, we first introduce the experimental setup and implementation details of the HopFIR networks. We then present our experimental results and comparisons with state-of-the-art methods.
Finally, we conduct several ablation studies of the proposed architecture.

\subsection{Datasets and Evaluation Protocols}
\textbf{Datasets.}  Human3.6M~\cite{ionescu2013h36m} is currently the largest publicly available dataset for 3D human pose estimation, with 3.6 million video frames. It captures accurate 3D human joint positions from four camera viewpoints and records 11 subjects performing 15 assigned actions. Following previous works~\cite{zou2021modulated, xu2021graph, zhao2019semantic}, we train our model on five subjects (S1, S5, S6, S7, S8) and test it on two subjects (S9, S11). In contrast to Human3.6M, MPI-INF-3DHP~\cite{mehta20173dhp} includes complex outdoor scenes, which are commonly used to evaluate the generalizability of proposed methods. Accordingly, we use the test set of MPI-INF-3DHP to verify the generalizability of our model.

\textbf{Evaluation Protocols.}  For Human3.6M~\cite{ionescu2013h36m}, we evaluate our model on two standard evaluation protocols: the mean per-joint position error (MPJPE) and the mean per-joint position error after Procrustes alignment (P-MPJPE). These are referred to Protocol \#1 and Protocol \#2, respectively.
The MPJPE and P-MPJPE are given in millimeters. 
For MPI-INF-3DHP~\cite{mehta20173dhp}, we follow previous works~\cite{ci2019LCN, zhao2019semantic, zou2021modulated} by reporting the percentage of correct keypoints (PCK) with a threshold of 150 mm and the area under the curve (AUC) for a range of PCK thresholds.

\subsection{Implementation Details}

Following previous work~\cite{pavllo20193d}, we obtain the detected 2D poses using the cascaded pyramid network(CPN)~\cite{chen2018cpn}.
We do not use data augmentation during training and testing with the 2D ground truth input to verify the efficacy of our model.
We adopt Adam~\cite{kingma2014adam} optimizer and all experiments are conducted on a single NVIDIA RTX 3090 GPU.
3D pose regression from 2D detections is more challenging than that from 2D ground truth because the former needs to deal with some extra uncertainty in the 2D space.
To manage this uncertainty, we set different configurations for them.
In experiments with 2D ground truth as the input, we train the HopFIR networks with an initial learning rate of 0.001, a decay factor of 0.90 per 4 epochs, a batch size of 64, channels of 128, and PReLU activation~\cite{he2015prelu}. When using detected 2D poses, we train the HopFIR networks with the initial learning rate of 0.006, a decay factor of 0.95 per 4 epochs (but 0.2 for the first 4 epochs), a batch size of 256, 256 channels, and LeaklyReLU activation~\cite{maas2013leakyrelu}.
To avoid overfitting, we apply Dropout\cite{JMLR:v15:dropout} with a dropout rate of 0.5.

\subsection{Comparison with State-of-the-art}

We compare the performance of HopFIR with some SOTA methods on Human3.6M under Protocol \#1 and Protocol \#2, with the results shown in Table~\ref{table:2D}. 
Our method reaches an MPJPE of 48.50 mm and outperforms the best of the existing approaches~\cite{zou2021modulated} on all 15 actions.
Given the uncertainty of 2D detections, we also investigate the capability of HopFIR networks using ground truth 2D key points as input. As shown in Table~\ref{table:GT1}, HopFIR obtains surprisingly better performance when given precise 2D joint information and produces SOTA results, which verifies its effectiveness.

\subsection{Ablation Study}

We conduct a comprehensive ablation study on Human3.6M to validate the individual effectiveness of each component of the proposed HopFIR architecture under controlled settings.
We follow previous works ~\cite{liu2020comprehensive, xu2021graph, zou2021modulated} conducting the ablation experiments using GT as inputs to avoid the influence of the 2D pose detector. 

\textbf{Effectiveness of Different Modules.}  
We separately verify the effectiveness of the HGF module and IJR module and conduct experiments on first-order neighbors, removing all modules and the multi-hop mechanism.
Note that GCN and HopGCN refer to applying only vanilla GCN (Eq.~\ref{e1}) and HopGCN (Eq.~\ref{e5}) to Fig.~\ref{fig:framework}, respectively.
The results in Table~\ref{table:layer} show that each module improves the performance over the GCN-only approach, and coupling the two layers to form a HopFIR block achieves further performance improvement. 
The HopFIR networks reduce the MPJPE to 32.67 mm, which represents a 7.2\% improvement over GraFormer~\cite{zhao2022graformer}.
By reducing feature channels in HopGCN \& IJR
and HopGCN \& HGF, we decrease parameters to 0.57M
and 0.48M. However, the models still achieved errors of
36.69mm and 36.01mm, compared to the HopGCN with
39.15mm error and 0.59M parameters, which should be attributed to the structure design other than the model size.
Moreover, we reduce the HopFIR network parameters by changing the channels to 64 and 32, respectively, which are also superior to SOTA.

\begin{table}[t]
    \begin{center}
    \setlength{\extrarowheight}{0pt}
    \setlength{\aboverulesep}{0pt}
    \setlength{\belowrulesep}{0pt}
    \resizebox{75mm}{!}{
    \begin{tabular}{l|c|c|c|c}
        \toprule
        Method & Channels & Params & MPJPE & P-MPJPE \\
        \midrule\midrule
        GCN &128 &0.36M &40.63 &31.65\\ 
        HopGCN &128 &0.59M &39.15 &31.40\\ 
        \midrule
        HopGCN \& IJR & 128 &2.15M &36.62 &29.23\\ 
        HopGCN \& HGF & 128 & 1.05M &35.19 &28.81 \\
        HopGCN \& IJR & 64 & 0.57M &36.69 &29.68\\ 
        HopGCN \& HGF & 80 & 0.48M &36.01 &29.58 \\
        \midrule
        HopGCN \& HopFIR & 128 & 2.15M &\textbf{32.67} &\textbf{26.20} \\ 
        HopGCN \& HopFIR & 64 & 0.54M & 33.52& 27.37\\ 
        HopGCN \& HopFIR & 32 & 0.14M & 35.19& 28.71\\ 
        \bottomrule
    \end{tabular}
    }
    \end{center}
    \caption{Ablation experiments on the proposed modules.
    }
    \label{table:layer}
\end{table}

\begin{figure}[t]
    \begin{center}
    \includegraphics[width=\linewidth]{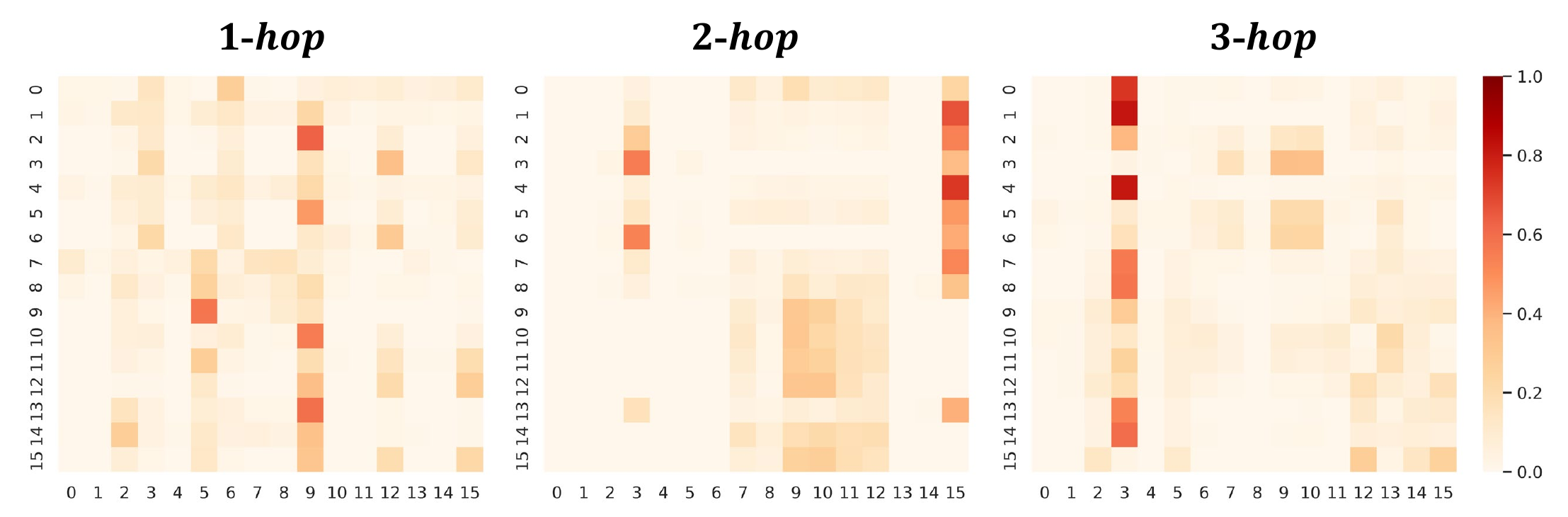}
    \end{center}
    \caption{
    Attention weight of the $j$-th $k$-hop for the $i$-th joint, deeper color indicates higher correlation.
    $i$-th row and $j$-th col represent $i$-th joint and $k$-hop of $j$-th joint, respectively.
    }
    \label{fig:weightgroup}
\end{figure}

\textbf{Effectiveness of the HA Layer.}  
In sec.~\ref{subsec:HGF}, we introduce $k$-hop groups to discover latent joint interactions in human joint synergies. 
The attention matrices in Fig.~\ref{fig:weightgroup} show the latent joint synergies captured by HopFIR, in which each weight of row $i$ indicates a discovered latent group for the corresponding joint $i$.
To verify the effectiveness of HA layer, we explore the correlation between all joints with a transformer encoder instead of HA layer, which explores the correlation between individual nodes but ignores the synergy between groups of joints in the human body and obtained 36.21 mm error. Moreover, we remove the human body prior by using random graph instead of skeleton graph without changing the HopFIR architecture and reached 34.68 mm error, suggesting that latent group correlations can be explored by $k$-hop groups, but group correlations underlying joint synergies can be better explored based on the human body prior.

In Table~\ref{table:HA_attn}, we show the experimental results of three different ways to design the HA layer. HSS is the method selected in this paper, where H, S, and S are $Q$, $K$, and $V$, and H and S represent the node feature and the $k$-hop group feature, respectively. 
As we do not follow~\cite{vaswani2017attention} in applying a linear transformation of $Q$, $K$, and $V$,
we also show the result of such a linear transformation,
which is denoted as HA+W. 
The experiment results show that HSS similarity achieves better performance in the HopFIR architecture, but one can choose the type of similarity according to the network property. 

We further insert the HA layer into SOTA pose estimation methods, namely SemGCN~\cite{zhao2019semantic} and Modulated GCN~\cite{zou2021modulated}, to investigate its generalizability. No changes are made to their source code, with the HA layer inserted before the information aggregation stage of these methods.
The experimental results in Table~\ref{table:HAmodule} show that the HA layer improves these previous SOTA networks to a large degree; especially, the MPJPE of SemGCN~\cite{zhao2019semantic}, is reduced from 42.14 mm to 38.41 mm, representing an 8.9\% improvement. 
Moreover, HA layer with linear transformation (W) makes learning more stable,
so We test HA and HA+W in SemGCN and MGCN,
respectively, to show both of them are effective. 
For a fair comparison of parameters, we tested MLP and MHSA instead of HA, both of them with 1.03M parameters, and achieved 39.12mm and 39.22mm errors, respectively.
More details can be found in the supplementary material.
Experiments on the above methods of aggregating first-order neighbor information demonstrate the effectiveness of HA and also indicate that latent joint grouping can recognize the human joint synergies.

\begin{table}[t]
    \small
    \begin{center}
    \setlength{\extrarowheight}{0pt}
    \setlength{\aboverulesep}{0pt}
    \setlength{\belowrulesep}{0pt}
    \resizebox{75mm}{!}{
    \begin{tabular}{l|ccc||ccc}
        \toprule
        \multirow{2}{*}{Attention} &\multicolumn{3}{c||}{HA} &\multicolumn{3}{c}{HA+W}\\
        \cline{2-2}\cline{3-3}\cline{4-4}\cline{5-5}\cline{6-6}\cline{7-7}
        &HSS &SSS &SHH &HSS &SSS &SHH\\
        \hhline{-|---||---}
        Params  &\multicolumn{3}{c||}{2.15M} &\multicolumn{3}{c}{2.50M}\\ 
        \hhline{-|---||---}
        MPJPE   &\textbf{32.67} &34.18 & 34.28 & \textbf{33.29} & 33.53 & 33.90\\ 
        P-MPJPE &\textbf{26.20} &27.70 & 27.71 & \textbf{27.16} & 27.40 & 27.41\\ 
        \bottomrule
    \end{tabular}
    }
    \end{center}
    \caption{Quantitative comparison of HA layers with different similarity computing approaches.
    }
    \label{table:HA_attn}
\end{table}

\begin{table}[t]
    \begin{center}
    \setlength{\extrarowheight}{0pt}
    \setlength{\aboverulesep}{0pt}
    \setlength{\belowrulesep}{0pt}
    \resizebox{83mm}{!}{
    \begin{tabular}{l|c|c|c|c|c}
        \toprule
        \multicolumn{2}{l|}{Method} & Channels & Params & MPJPE & P-MPJPE \\
        \midrule\midrule
        \multicolumn{2}{l|}{SemGCN~\cite{zhao2019semantic}} & 128 & 0.27M & 42.14 & 33.53 \\ 
        \multicolumn{2}{l|}{SemGCN + HA(HSS)} & 128 & 0.49M &\textbf{38.41} & \textbf{30.56} \\ 
        \multicolumn{2}{l|}{SemGCN + HA(SSS)} & 128 & 0.49M & 41.30 & 33.07 \\ 
        \multicolumn{2}{l|}{SemGCN + HA(SHH)} & 128 & 0.49M & 38.81 & 31.05 \\ 
        \midrule
        \multicolumn{2}{l|}{SemGCN~\cite{zhao2019semantic} w/ Non-local~\cite{wang2018nonlocal}} & 128 & 0.43M & 40.78 & 31.46 \\ 
        \multicolumn{2}{l|}{SemGCN w/ Non-local +HA(HSS)} & 128 & 0.66M & 38.03 & 30.50 \\ 
        \multicolumn{2}{l|}{SemGCN w/ Non-local +HA(SSS)} &128 &0.66M &\textbf{37.75} &30.17 \\ 
        \multicolumn{2}{l|}{SemGCN w/ Non-local +HA(SHH)} &128 &0.66M &37.94 &\textbf{29.71} \\ 
        \midrule
        \multicolumn{2}{l|}{Modulated GCN~\cite{zou2021modulated}} & 128 & 0.29M & 38.25 & 30.06 \\ 
        \multicolumn{2}{l|}{Modulated GCN +HA(HSS)+W} & 128 &0.96M &36.54 & 29.09\\ 
        \multicolumn{2}{l|}{Modulated GCN +HA(SSS)+W} & 128 &0.96M &\textbf{36.14} &\textbf{29.02}\\ 
        \multicolumn{2}{l|}{Modulated GCN +HA(SHH)+W} & 128 &0.96M &37.38 &30.02\\ 
        \bottomrule
    \end{tabular}
    }
    \end{center}
    \caption{Comparison of the improved performance of proposed HA layer added on different methods. We test on two GCN-based methods: SemGCN~\cite{zhao2019semantic} and MGCN~\cite{zou2021modulated}.
    }
    \label{table:HAmodule}
\end{table}

\begin{figure}[t]
  \begin{center}
  \includegraphics[width=0.8\linewidth]{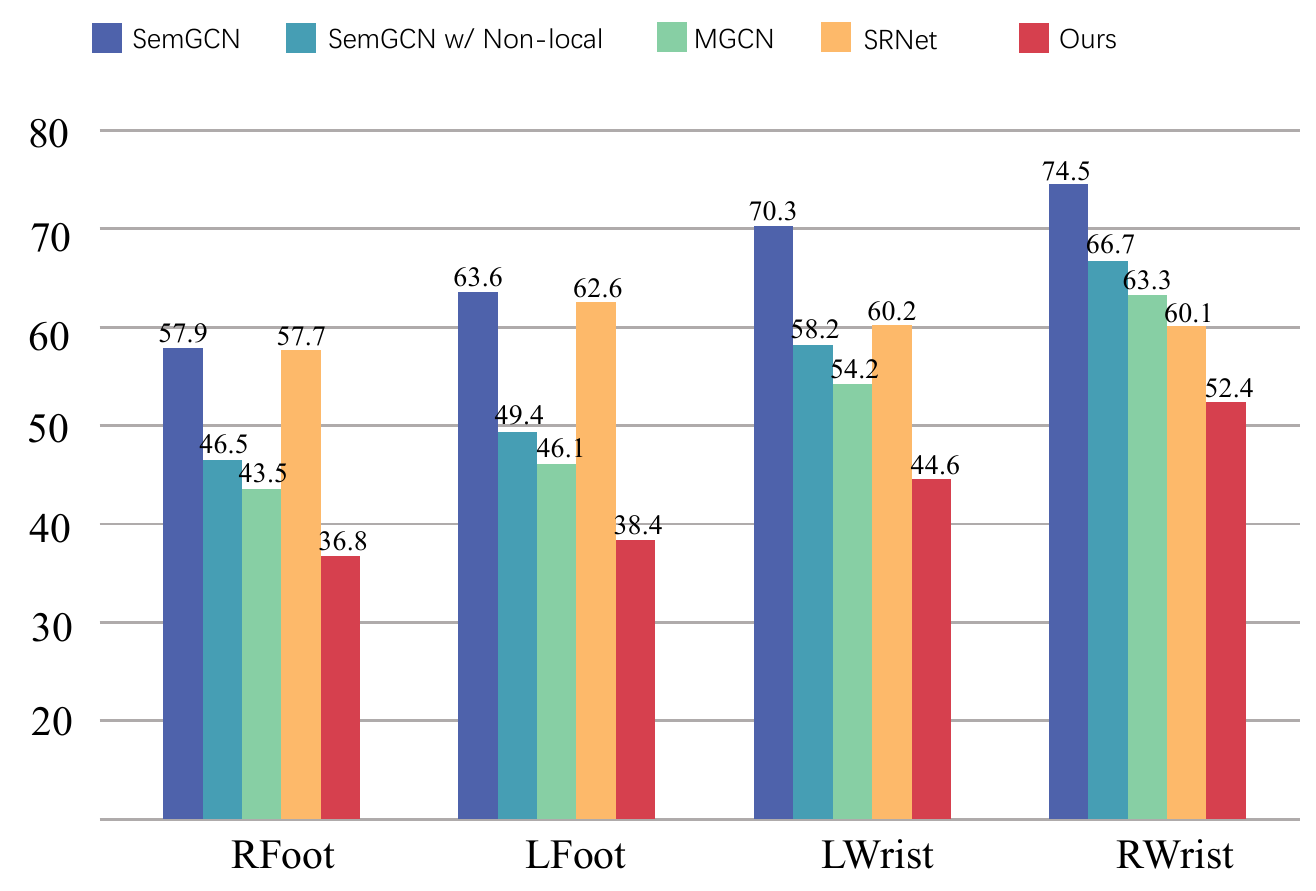}
  \end{center}
  \caption{Comparison of
   MPJPE for peripheral joints on the test set of the Human3.6M. R and L denote right and left, respectively.}
  \label{fig:joint_error}
\end{figure}

\begin{table}[t]
    \begin{center}
    \setlength{\extrarowheight}{0pt}
    \setlength{\aboverulesep}{0pt}
    \setlength{\belowrulesep}{0pt}
    \resizebox{65mm}{!}{
    \begin{tabular}{c|c|c|c|c}
        \toprule
        Num-$k$ & Channels & Params & MPJPE & P-MPJPE \\
        \midrule\midrule
        1 & 128 & 1.88M & 35.88 & 28.76 \\ 
        2 & 128 & 2.03M & 34.96 & 27.74 \\ 
        3 & 128 & 2.15M &\textbf{32.67} &\textbf{26.20} \\ 
        4 & 128 & 2.27M & 35.58 & 28.35  \\ 
        \midrule
        Num-Block & Channels & Params & MPJPE & P-MPJPE \\
        \midrule\midrule
        1 & 128 &0.80M &37.33 &30.71\\ 
        2 & 128 &1.47M &34.21 &27.63 \\ 
        3 & 128 &2.15M &\textbf{32.67} &\textbf{26.20}\\ 
        4 & 128 &2.82M &33.84 &27.63\\ 
        \bottomrule
    \end{tabular}
    }
    \end{center}
    \caption{Ablation study for number of $k$-hop and designed blocks. The units of MPJPE and P-MPJPE are millimeters (mm).}
    \label{table:hopnum}
\end{table}

\begin{table}[t]
    \begin{center}
    \setlength{\extrarowheight}{0pt}
    \setlength{\aboverulesep}{0pt}
    \setlength{\belowrulesep}{0pt}
    \resizebox{65mm}{!}{
    \begin{tabular}{c|c|c|c|c}
        \toprule
        Block & Channels & Params & MPJPE & P-MPJPE \\
        \midrule\midrule
        (H)(I) & 128 &1.78M &34.43 &27.79 \\ 
        (I)(H) & 128 &1.78M &34.13 &27.56 \\ 
        (H)(I)(H) & 128 & 2.15M &35.08 &27.98 \\ 
        (I)(H)(H) & 128 &2.15M &34.20 &27.60  \\ 
        (H)(H)(I) & 128 &2.15M &\textbf{32.67} &\textbf{26.20}\\ 
        \bottomrule
    \end{tabular}
    }
    \end{center}
    \caption{Ablation study for arrangements of the designed block. (I) and (H) denote IJR module and HGF module, respectively.}
    \label{table:block}
\end{table}

\begin{table}[ht]
    \begin{center}
    \setlength{\extrarowheight}{0pt}
    \setlength{\aboverulesep}{0pt}
    \setlength{\belowrulesep}{0pt}
    \resizebox{70mm}{!}{
    \begin{tabular}{l|c|c|c}
        \toprule
        &MPJPE
        &P-MPJPE 
        &MPJVE\\
        \midrule
        \midrule
        PoseFormer~\cite{zheng2021poseformer} ($T$=81) &44.3 &36.5 &3.1\\
        MixSTE~\cite{zhang2022mixste} ($T$=243) &\textbf{40.9}&\underline{32.6}&\underline{2.3}\\
        MHFormer~\cite{li2022mhformer} ($T$=351)&43.0&- &- \\
        P-STMO~\cite{shan2022p} ($T$=243)&42.1&34.4&-  \\
        \midrule
        Ours ($T$=243)&\underline{41.1} &\textbf{32.5}&\textbf{2.1}\\
        \bottomrule
    \end{tabular}
    }
    \end{center}
    \caption{Quantitative comparision on Human3.6M with detected 2D pose (CPN) in video. $T$ denotes the number of input frames.}
    \label{tab: CPN_temporal}
\end{table}

\begin{table}[t]
    \begin{center}
    \setlength{\extrarowheight}{0pt}
    \setlength{\aboverulesep}{0pt}
    \setlength{\belowrulesep}{0pt}
    \resizebox{80mm}{!}{
    \begin{tabular}{l|c|c|c|c|c}
        \toprule
        \multirow{2}{*}{Methods}&\multicolumn{4}{c|}{PCK} & \multirow{2}{*}{AUC} \\
        \cline{2-2}\cline{3-3}\cline{4-4}\cline{5-5}
        & GS & no GS & Outdoor & All\\
        \midrule\midrule
        Martinez et al.~\cite{martinez2017simple} & 49.8 & 42.5 & 31.2 & 42.5 & 17.0\\ 
        Ci et al.~\cite{ci2019LCN} & 74.8 & 70.8 & 77.3 & 74.0 & 36.7\\ 
        zeng et al.~\cite{zeng2020srnet} & - & - & 80.3 & 77.6 & 43.8\\ 
        Li et al.~\cite{li2019generating} & 70.1 & 68.2 & 66.6 & 66.9 & -\\ 
        Zhao et al.~\cite{zhao2022graformer}&80.1 &77.9 &74.1 &79.0 & 43.8\\ 
        Liu et al.~\cite{liu2020comprehensive} (weight unsharing)& 77.6 & 80.5 & 80.1 & 79.3 & 47.6\\ 
        Xu et al.~\cite{xu2021graph} & 81.5 & 81.7 & 75.2 & 80.1 & 45.8\\
        Nie et al.~\cite{nie2023lifting} & - & - & - & 83.5 & 45.9\\
        Zou et al.~\cite{zou2021modulated} & 86.4 & \textbf{86.0} &85.7 & 86.1 & 53.7\\ 
        \midrule
        Ours & \textbf{89.1} & 85.9 & \textbf{85.9} & \textbf{87.2} & \textbf{57.0}\\ 
        \bottomrule
    \end{tabular}
    }
    \end{center}
    \caption{Quantitative comparisons on the MPI-INF-3DHP test set. GS denotes green screen.
    }
    \label{table:mpi3d}
\end{table}

\begin{figure}[t]
  \begin{center}
  \includegraphics[width=1\linewidth]{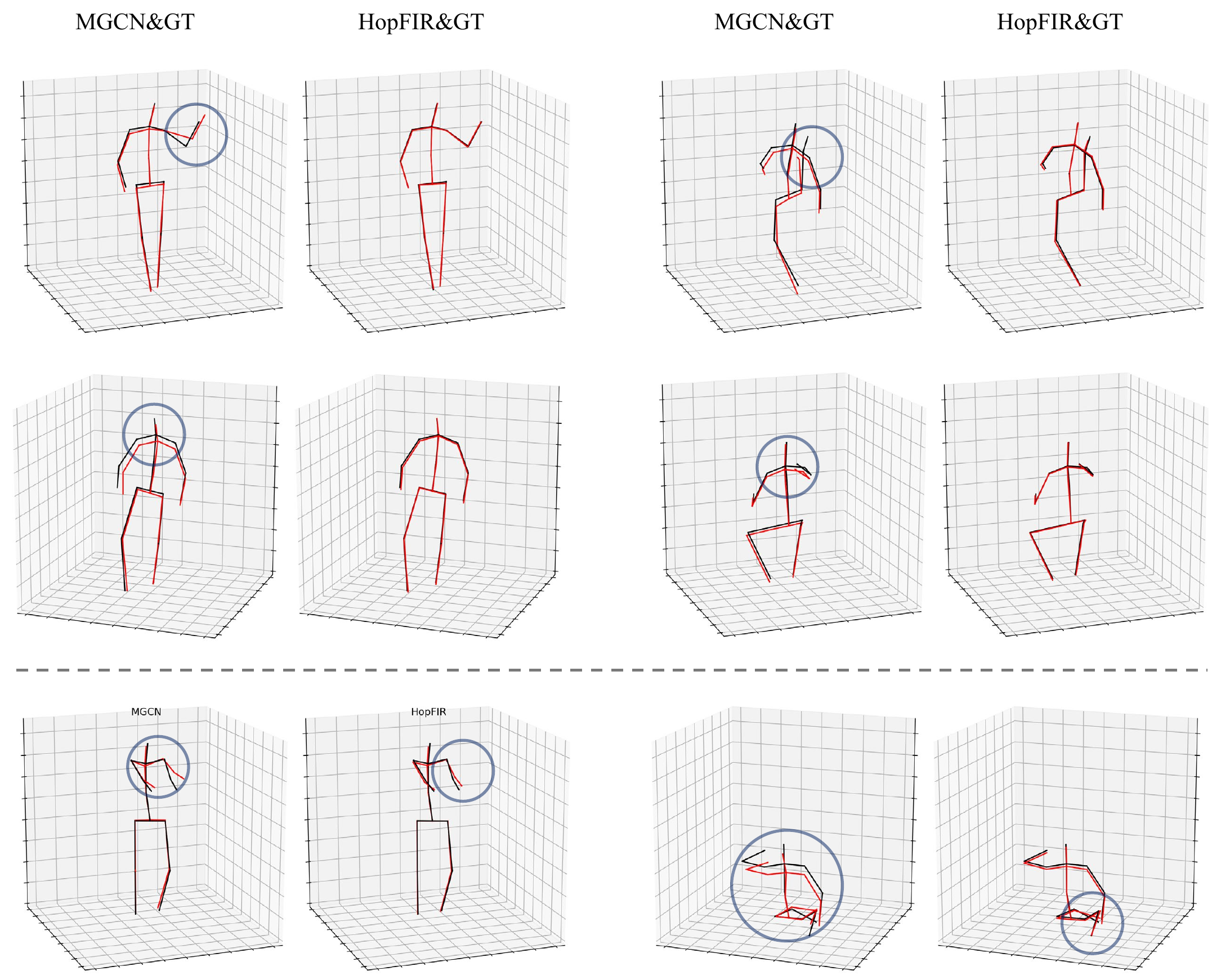}
  \end{center}
  \caption{
  Qualitative visual results for HopFIR and MGCN~\cite{zou2021modulated} on the Human3.6M. The black lines are the ground truth (GT) and the red lines are the predictions of HopFIR and MGCN. Wrong predictions are circled.
  The bottom row shows our failure case.
  }
  \label{fig:qualitative}
\end{figure}

\textbf{Error on Peripheral Joints.}  
We report the regression accuracy for the peripheral joints (wrists and feet) in Fig.~\ref{fig:joint_error} in comparison with some previously proposed methods~\cite{zou2021modulated, zhao2019semantic,zeng2020srnet}. The HopFIR network with IJR modules outperforms SOTA methods on the right foot (RFoot) by 6.7 mm, left foot (LFoot) by 7.7 mm, left wrist (LWrist) by 9.6 mm, and right wrist (RWrist) by 7.7 mm.
The experimental results verify that the intragroup joint attention within each limb group strengthens the capabilities of the HopFIR.

\textbf{Different Numbers of $k$-Hops.}  
As HopFIR is designed to extract the correlation between feature groups, we set different $k$ values to discover various latent connections underlying the human joint synergies, with the results shown at the top of Table~\ref{table:hopnum}.
The MPJPE gradually decreases as the number of hops increases, and reaches the best performance at 3 hops. Therefore, the optimal number of hops for 3D HPE is 3, which entails that we obtain $16\times3$ groups from the skeleton graph. Each of the 16 groups corresponds to a potential correlation among coupled nodes at different distances, and three hops is sufficient to recognize the joint synergies.

\textbf{Arrangement of the Designed Block.}  To investigate the optimal structure of the designed block, experiments are performed with various block numbers and various combinations of HGF and IJR modules. 
As shown at the bottom of Table~\ref{table:hopnum}, the error gradually decreases as the number of blocks increases, until the best performance is achieved at 3 blocks.
As shown in Table~\ref{table:block}, the (H)(H)(I) arrangement achieves the optimal results by reducing the error to 32.67 mm.
HGF treats each hop as a group and applies a hop-wise attention mechanism to these groups to discover latent joint synergy. IJR utilizes the limb prior for peripheral joint refinement. Thus, (H)(H)(I) first integrates the complete joint information and then refines it by IJR. While (I)(H)(H) reverses this procedure resulting in the insufficient utilization of joint information.
That the results are superior to GraFormer~\cite{zhao2022graformer} in all cases except for the single-block model indicates that the HopFIR has significant human pose representation capabilities.

\textbf{Extend to Temporal Domain.}
Without a specific design to integrate temporal information, we extend to the temporal domain by adding two TEs after each block and replacing the HG module with a linear layer, and achieved competitive results as shown in table~\ref{tab: CPN_temporal}. 

\textbf{Cross-Dataset Results on MPI-INF-3DHP.}  
Table~\ref{table:mpi3d} further compares HopFIR with previous methods on cross-dataset scenarios to validate its generalizability. For these experiments, we train our model on the Human3.6M dataset and test it on the test set of the MPI-INF-3DHP dataset. 
The results show that our approach obtains better results than other methods, which verifies the generalizability of our approach to unseen scenarios.
\textbf{Qualitative Results.}  In Fig.~\ref{fig:qualitative}, we show the visual results on Human3.6M in the world space. 
The bottom of the figure shows some failure cases of HopFIR, which predict some wrong joint positions. 
The figure shows that HopFIR is able to predict 3D joint positions more accurately, even for poses that cause difficulties for MGCN. 

\section{Conclusions}
We present the Hop-wise GraphFormer with Intragroup Joint Refinement (HopFIR) as a novel architecture for 3D human pose estimation. The proposed architecture mainly comprises the HGF and IJR modules. The HGF module improves on the GCN-based pose estimation networks by grouping the joints by $k$-hop neighborhood and capturing the potential joint correlations in the different joint synergies. Because the peripheral joints strongly interact with intra-limb joints, the proposed IJR module applies intragroup attention to refine the peripheral joint features through the associated limb. The proposed method achieves new state-of-the-art results while maintaining a modest model size.

\section*{Acknowledgement} 
This work was supported in part by the National Key Research and Development Program of China (Grant No. 2021YFC0122602), 
in part by the Joint Funds Program of the National Natural Science Foundation of China (Grant No. U21A20517), 
in part by the Basic Science Centre Program of National Natural Science Foundation of China (Grant No. 72188101).

{\small
\bibliographystyle{ieee_fullname}
\bibliography{egbib}
}

\end{document}